%% file: main.tex
\documentclass[10pt,twocolumn,letterpaper]{article}

\usepackage{cvpr}
\usepackage{times}
\usepackage{epsfig}
\usepackage{graphicx}
\usepackage{amsmath}
\usepackage{amssymb}
\usepackage[boxed]{algorithm2e}
\usepackage{sg-macros}
\usepackage{subfig}

\SetKwInput{KwInput}{Input}
\SetKwInput{KwOutput}{Output}
\definecolor{grey}{rgb}{0.5, 0.5, 0.5}

\SetCommentSty{commentfont}

\newcommand{\background}{\diamond}

\usepackage[pagebackref=true,breaklinks=true,letterpaper=true,colorlinks,bookmarks=false]{hyperref}

\cvprfinalcopy 

\def\cvprPaperID{820} 

\ifcvprfinal\fi
\begin{document}

\title{Unsupervised Human Action Detection by Action Matching}


\author{Basura~Fernando\footnotemark[1]\hspace{0.25em}\quad{}Sareh~Shirazi\footnotemark[2]\hspace{0.25em}\quad{}Stephen Gould\footnotemark[1]\\
\begin{tabular}{cc}
    \footnotemark[1] The Australian National University &\footnotemark[2] Queensland University of Technology\\
    {\tt\small firstname.lastname@anu.edu.au} & {\tt\small s.shirazi@qut.edu.au}
    \tabularnewline    
\end{tabular}}

\maketitle

\begin{abstract}
  \input{abstract}
\end{abstract}

\section{Introduction}
\input{intro}

\section{Related work}
\input{related}

\section{Unsupervised action detection}
\input{method}

\section{Experiments}
\input{experiments}

\section{Conclusion}
\input{conclusion}

\noindent
\textbf{Acknowledgement} : {\small This research was conducted by the Australian Research Council Centre of Excellence for Robotic Vision (project number CE140100016) and was undertaken on the NCI National Facility in Canberra, Australia, which is supported by the Australian Commonwealth Government.}

{\small
\bibliographystyle{ieee}
\bibliography{videoalignment}
}

\end{document}

%% file: abstract.tex
We propose a new task of unsupervised action detection by action matching.
Given two long videos, the objective is to temporally detect all pairs of matching video segments. 
A pair of video segments are matched if they share the same human action. 
The task is category independent---it does not matter what action is being performed---and no supervision is used to discover such video segments. 
Unsupervised action detection by action matching allows us to align videos in a meaningful manner.
As such, it can be used to discover new action categories or as an action proposal technique within, say, an action detection pipeline.
Moreover, it is a useful pre-processing step for generating video highlights, e.g., from sports videos.

We present an effective and efficient method for unsupervised action detection. We use an unsupervised temporal encoding method and exploit the temporal consistency in human actions to obtain candidate action segments. We evaluate our method on this challenging task using three activity recognition benchmarks, namely, the MPII Cooking activities dataset, the THUMOS15 action detection benchmark and a new dataset called the IKEA dataset. On the MPII Cooking dataset we detect action segments with a precision of 21.6\% and recall of 11.7\% over 946 long video pairs and over 5000 ground truth action segments. Similarly, on THUMOS dataset we obtain 18.4\% precision and 25.1\% recall over 5094 ground truth action segment pairs.

%% file: intro.tex
Recognizing human activities in unconstrained videos is important for many applications including human computer interaction, human robots interaction, sports video analysis, video retrieval, storyline reconstruction and for many other video analysis tasks~\cite{Herath2017,Poppe2010}.
However, it is hard to define what a \emph{human action is}. 
In the current literature, human actions are defined based on tasks such as cutting, washing~\cite{Rohrbach2012}, based on specificity and regularity of human motion such as running, walking, hand waving~\cite{Schuldt2004}, or based on sports activities such as weight lifting, skying or cricket bowling~\cite{Soomro2012}. 
Moreover, current methods in human action recognition require a lot of supervised data~\cite{Karpathy2014}. 
Human action detection is the task of temporally localizing a human action within a long video~\cite{Rohrbach2012}. 
Obtaining ground truth labels for human actions in video collections is costly and consequently annotated large high quality video datasets are hard to come by. Unlike action classification, which just requires a single label for the entire video sequence, to create an action detection dataset the annotator must watch the entire video and mark the beginning and end of each human action.
Such manual annotations could be wrong, subjective and highly ambiguous.
But action detection methods require a lot of annotations to supervise training of action detectors. As such, a more efficient method is needed.

\begin{figure}[t]
\begin{center}
  \includegraphics[width=0.99\linewidth]{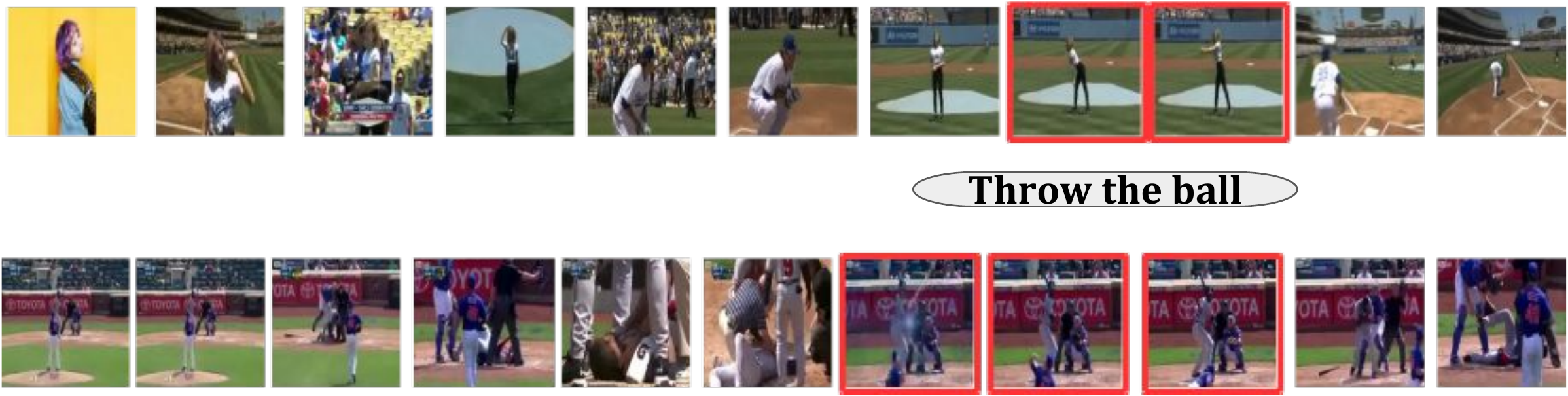}
\end{center}
\caption{Unsupervised action detection by action matching. Two videos share a common human action \emph{throw the ball}. Objective is to temporally localize the common human action units (segments) within a pair of videos.}
\label{fig:sample}
\end{figure}

In this paper, we present a method to discover common action segments from a pair of videos based on human action matching in \emph{unsupervised manner}. We call this novel task \emph{unsupervised action detection by action matching}. We match segments of one video with the other one such that matched segments are from the same human action category. The method must recognize similar human action segments while temporally localizing the actions (i.e.,~detection) without using any external information (see Figure~\ref{fig:sample}). As the task is unsupervised, our method does not know the type of detected action, only that one is occurring. 
This task extends the ASLAN~\cite{Kliper-Gross2012} challenge where the task was to predict whether a pair of videos contains the same action or not.
However, in the proposed task, the pair of videos may contain a series of human actions and a large number of matched action segments.

Unsupervised action detection by action matching is useful for many applications. Obviously, the output of this task can be used to discover human action categories in unsupervised manner, which could be subsequently labeled.
For example, we can cluster the large number of matched human action segments from unsupervised action detection to discover human action categories similar to unsupervised object discovery in static images~\cite{Tuytelaars2010}.
The proposed task is also useful for early human action and activity prediction~\cite{Ryoo2011}. 
Imagine a robot that has access to a large video archive containing many human actions and tasks (such as cooking a meal, fixing a table, and cleaning a garage).
Now whenever the robot sees a particular sequence of human actions in a live video stream, it is able to align common actions in the live stream with the videos in the archive.
This would allow the robot to temporally localize the human activities without supervision. 
Moreover, the robot could anticipate future human actions by aligning what it has seen with the archive without any annotations. 
Another application of the proposed task is in video highlights generation of, say, sports activities without human intervention~\cite{Hanjalic2005}. 
Imagine a system with access to an archive of sports highlights videos. 
Given a full sports coverage video, the system can generate new highlights by aligning the highlight videos with full coverage video through detection and alignment of human actions. 
Since highlights only capture interesting events, we can exploit such prior information to generate new video sports highlights. Yet another application for the proposed task is to generate candidate temporal action proposals similar to object proposal methods that are currently popular for object detection in static images~\cite{Carreira2012}.

In this paper we propose a novel unsupervised action detection task and propose a very effective yet simple method to solve the problem by human action matching. The related task of action classification is a well studied with much progress being made over the last decade~\cite{Feichtenhofer2016,Jaincvpr15,Lan2015a,Tran2015,Wang2015,Simonyan2014,PengECCV2014}. 
However, action detection is a relatively new task that has shown great promise in recent years~\cite{Rohrbach2012,Ni2016,Singh2016,Yeung2016,Richard2016}. Action detection is challenging as the duration of the action is varied and unknown, and the background context can easily confuse the action classifiers. Furthermore, video computation is expensive so very effective temporal encoding methods are needed. In supervised action detection, one can rely on a large source of annotated videos of human actions to obtain discriminative temporal encodings using discriminative sequence encoders~\cite{Donahue2015}. 
As the proposed task is unsupervised, however, we must rely on unsupervised video temporal dynamic encoding techniques such as unsupervised LSTMs~\cite{Srivastava2015} or rank pooling~\cite{Fernando2016}. 
Due to its demonstrated effectiveness, we use a variant of rank pooling with the addition of temporal consistency for the task of \emph{unsupervised action detection}. The naive approach of simply matching all possible video segments will not produce good results due to large number of false positives and deficiencies in the matching function. At the same time such an approach would not be able to take advantage of temporal consistency and smoothness in the execution of human actions. Moreover, it is highly inefficient. To overcome these drawbacks, we present a simple yet effective algorithm that exploits the temporal smoothness and consistency of human action evolution. Results of our method are reported on three activity recognition benchmarks.

%% file: related.tex
Video alignment and video synchronization is related to the proposed task. However, except one instance we could not find any prior work in video synchronization or alignment that uses semantics such as human actions to align videos \cite{Ukrainitz2006}. 
Most prior related work align pair of videos both in temporal and spatial domain without considering the action semantics~\cite{Diego2011} or human action dynamics. This is mostly done by sequence to sequence matching via frame correspondences. Frames content should be matched and correspondences should be found. Most prior work exploit geometric and photometric properties of the two scenes to find the correspondences in space and time~\cite{Diego2011}. For example, in Diego~\etal~\cite{Diego2011} the paper assumes that the motion in two videos are somewhat similar and there is some overlap between field of view of cameras. Similarly, in Ukrainitz and Irani~\cite{Ukrainitz2006} alignment is performed in space and time by maximizing the local space-time correlations directly using the pixel intensity information. They seek a transformation that minimizes the spatial-temporal displacement of near identical pair of videos. Cross-view action recognition by exploiting the self similarity of videos is also somewhat related to our work~\cite{Junejo2008}. However, compared to other related methods, ours does not rely on geometric or photometric properties of pair of matching videos. We only rely on encoding of human motion dynamics. In contrast to these prior work we exploit the temporal consistency and smoothness of information evolution of human actions.
Furthermore, our task is an action detection task. To the best of our knowledge, unsupervised action detection is a novel task.

Supervised action detection is also related to ours~\cite{Rohrbach2012,Ni2016,Singh2016,Yeung2016,Richard2016}. Most of the progress in action detection is thanks to two main stream action detection datasets; the THUMOS challenge~\cite{THUMOS15} and the MPII cooking activity dataset~\cite{Rohrbach2012}. Rohrbach \etal~\cite{Rohrbach2012} perform action detection using dense trajectory features encoded with bag-of-words and then applying simple temporal pooling method such as sum-pooling followed by SVM classifiers. They use a sliding window method. Joint exploitation of geometrical contextual information among objects, human body parts, body poses is used for action detection using LSTM in Ni~\etal~\cite{Ni2016}. A method for fine-grained action detection in long video sequences based on a multi-stream bi-directional recurrent neural networks was presented in~Singh~\etal~\cite{Singh2016}. Reinforcement learning based action detection method that utilizes recurrent neural networks has also been studies~\cite{Yeung2016}. Different from all above methods, ours is an unsupervised action detection task where we have to temporally localize similar human actions in two long video sequences. 

Recent work on temporal action proposal is also related to our work~\cite{Escorcia2016}. However, most of these methods are supervised. Output of our method can be used for temporal action proposals in an unsupervised manner and is agnostic to the action category.

Our method is also related to rank-pooling based action recognition~\cite{bilen2016action,Bilen2016,Cherian2017,Fernando2016b,Fernando2015,Fernando2016,Fernando2016a}. Rank pooling method was first introduced to model the temporal evaluation of video sequences~\cite{Fernando2015,Fernando2016}. Furthermore, it was extended with hierarchical encoding~\cite{Fernando2016b}, end-to-end video representation learning~\cite{Fernando2016a}, and with subspaces~\cite{Cherian2017,Fernando2016}. Rank pooling principle was also used at input level~\cite{Bilen2016} or feature map level~\cite{bilen2016action}. To the best of our knowledge, we are the first to use rank-pooling based dynamic encoding for an action detection task.

%% file: method.tex
In this section we formalize the action matching problem (\S\ref{sec:problem}), provide an overview of our proposed solution (\S\ref{sec.overview}), discuss our method for sequence encoding (\S\ref{sec.enc}), and give details of how these come together to form a complete algorithm for \emph{unsupervised action detection by action matching} (\S\ref{sec.gram} and \ref{sec.consistency}). Finally, we present two strong baseline methods (\S~\ref{sec.baseline}) that we compare against in the experiments.


\subsection{Problem formulation}
\label{sec:problem}
Given a pair of video sequences $(X_a, X_b)$ where $X_a = \left< \bx_1^a, \bx_2^a, \ldots, \bx_n^a \right>$ and $X_b = \left< \bx_1^b, \bx_2^b, \ldots, \bx_m^b \right>$, we want to identify common human actions and localize them in each sequence. Many of the video frames may not belong to any human action class and we denote these by the special background label $\background$. Let us denote the set of human action categories that we care about by ${\cal Y}$. Then for each video $X = \left< \bx_1, \bx_2, \ldots, \bx_n \right>$, there is a corresponding label sequence $Y = \left< y_1, y_2, \ldots, y_n \right>$ where $y_t \in {\cal Y} \cup \{\background\}$ is the label for the $t$-th frame in the sequence. An action unit of video $X$ is a contiguous subsequence of $X$ that contains frames of only a single action class from the action label set ${\cal Y}$. An action unit $u$ is called maximal if frames adjacent to those in $u$ take a different action label from ${\cal Y} \cup \{\background\}$. An action unit $u_a$ from video $X_a$ is matched to action unit $u_b$ from video $X_b$ if the action class label of $u_a$ is same as action class label of $u_b$. Note that each action unit is valid if the union over intersection between an action unit and any ground truth is greater than some threshold (IoU = 0.5). The goal of our task is to find matching pairs of \emph{valid maximal action units} from a pair of arbitrary long videos.

\subsection{Overview of proposed solution}
\label{sec.overview}

Given a collection of videos, first we extract frame-level CNN features from them. With slight abuse of notation we denote the sequence of vectors for the $i$-th video as $X_i = \left< \bx_1^i, \bx_2^i, \ldots, \bx_n^i \right>$. We then sample subsequences of length $l_w$ and stride $l_s$ and apply temporal encoding to each subsequence. For every pair of videos from the collection we construct a similarity matrix between subsequences from the first and second video, respectively. Next, we exploit the temporal consistency of activities to obtain candidate action unit pairs. Last, we apply non-maximal suppression to remove redundant candidate action unit pairs.

\begin{figure}[t]
\begin{center}
  \includegraphics[width=0.6\linewidth]{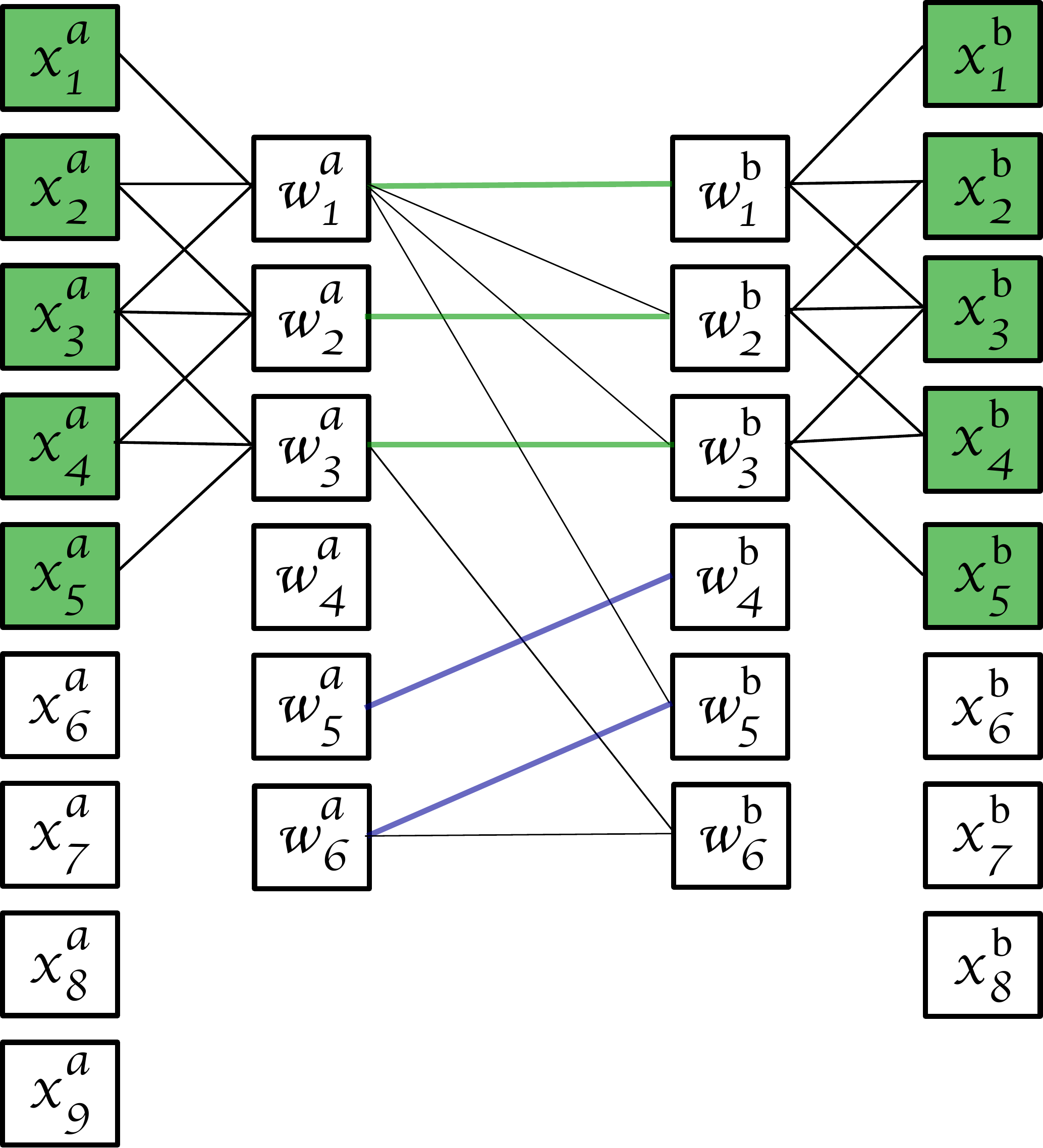}
\end{center}
\caption{Unsupervised action detection by action matching using temporally constant matching. Frames are denoted by $x_t^a$ for the first video and by $x_t^b$ for the second video. Then subsequences of size three are temporally encoded to obtain vectors $w_t^a$ and $w_t^b$. Afterwards, these temporal vectors are matched using bipartite graph matching. Matching temporal encodings from two videos are connected with an edge. Temporally consistent edges are shown in green and blue edges while isolated edges are shown in black. Each temporal encoding propagates the matches to frames as shown by the green coloured boxes. This way we can find the matching video segments that share similar temporal evolutions.}
\label{fig:bipartite}
\end{figure}

\subsection{Temporal encoding of video segments}
\label{sec.enc}

Temporal encoding takes an arbitrary long subsequence and represents it by a fixed length vector. Recently rank pooling was introduced as an effective and efficient method for temporal encoding of human actions~\cite{Fernando2016}. The first step applied by Fernando~\etal~\cite{Fernando2016} is to smooth the frame-level features by time varying means.

Given an input sequence $X = \left< \bx_1, \bx_2, \ldots, \bx_n \right>$, the time varying mean at frame $t$ is given by $\bm_t =\frac{1}{t} \sum_{\tau=1}^{t} \bx_\tau$. The method then normalizes the vector $\bm_t$ and only looks at the direction of the evolution of the mean. Let us denote this preprocessed sequence by $V = \left< \bv_1, \bv_2, \ldots, \bv_n \right>$, where each element $\bv_t$ is given by
\begin{align}
  \bv_t &= \frac{1}{\|\bm_t\|} \bm_t
\end{align}

Note that $\bv_t$ captures only the direction of the unit mean appearance vector at time $t$. Rank pooling then takes this pre-processed sequence $V$ and models the evolution of data over time using a linear ranking objective~\cite{Fernando2016} as follows:
\begin{align}
 \bw^\star \!&\in \argmin_{\bw} \!\left\{ \!\frac{1}{2}\|\bw\|^{2} + \frac{C}{2} \sum_{t=1}^J \Big[ |t - \bw^\top \bv_t| - \epsilon\Big]_{\geq 0}^2 \!\right\}
 \label{eq.svr}
\end{align}
As the parameter vector $\bw^\star$ models the evolution of appearance information within the video, it also captures the temporal structure that can be used to represent the dynamics effectively. Such representations are robust and efficient to compute with stand packages such as LibLinear~\cite{Fan2008}. In the original rank-pooling method~\cite{Fernando2016} non-linearity is introduced before the temporal encoding and after temporal encoding using a point-wise non-linear function $\Psi(\cdot)$, such as the signed square root. In summary, the steps of temporal encodings proposed by Fernando~\etal~\cite{Fernando2016} for supervised action classification is shown in the following equation:
\begin{equation}
\resizebox{0.47\textwidth}{!}{$X \overset{\bm_t = \frac{1}{t} \sum_{\tau=1}^{t} \bx_\tau}{\longmapsto} M
 \overset{\bv_t = \frac{\bm_t}{\|\bm_t\|} }{\longmapsto} V
 \overset{\Psi(\bv_t) } {\longmapsto} \widetilde{V}
 \overset{\Phi(\widetilde{V}) }{\longmapsto} \bw
 \overset{\Psi(\bw) }{\longmapsto} \widetilde{\bw}
 \overset{\frac{\widetilde{\bw}}{\|\widetilde{\bw}\|} }{\longmapsto} \widetilde{\bw^*}$}
 \label{eq.steps.rp}
\end{equation}

To simplify the pre-processing pipeline in this work we remove the non-linear transformations before and after rank pooling. Our pipeline for generating a temporal encoding of a video sequence is then:
\begin{equation}
X \overset{\bm_t}{\longmapsto} M
 \overset{\bv_t = \frac{\bm_t}{\|\bm_t\|} }{\longmapsto} V 
 \overset{\Phi({V}) }{\longmapsto} \bw 
 \overset{\frac{\bw}{\|\bw\|} }{\longmapsto} \bw^*
 \label{eq.steps.us}
\end{equation}

For long sequences the time varying mean smoothing applied by Fernando~\etal~\cite{Fernando2016} could be problematic in that smoothing is dominated by early frames. In this paper, we investigate an alternative scheme to obtain the smoothed vector $\bm_t$. Specifically, we investigate the following ARMA model in addition to the time varying mean,
\begin{align}
\bm_t &= \alpha \bm_{t-1} + (1  - \alpha) \bx_t
\end{align}
where we set $\bm_0$ to $\bx_1$.

\subsection{Temporal gram matrix construction}
\label{sec.gram}

Given a pair of long videos $X_a$ and $X_b$, we generate subsequences of length $l_w$ with a stride of $l_s$. Let the number of subsequences be denoted by $A$ and $B$ for $X_a$ and $X_b$, respectively. To avoid ambiguity in the sequel we refer to the subsequences of video frames as video segments. For each video segment we extract frame-wise features and apply rank pooling to obtain a fixed-length temporal encoding for each of the segments using the process explained in \eqnref{eq.steps.us}. The obtained ordered set of temporally rank pooled vectors is then $\left<\bw_1^a, \bw_2^a, \ldots, \bw_A^a\right>$ for video $X_a$ and $\left<\bw_1^b, \bw_2^b, \ldots, \bw_B^b\right>$ for video $X_b$. We can now construct a similarity matrix (Gram matrix) $G$ of size $A \times B$ by taking the inner-product between each pair of segments from videos $X_a$ and $X_b$ as
\begin{align}
 G_{i,j} &= \bw_i^a \cdot \bw_j^b,
 & \begin{matrix}
  \forall i \in \{1, \ldots, A\} \\
  \forall j \in \{1, \ldots, B\}
  \end{matrix}
\label{eq:gram}
\end{align}

\subsection{Exploiting the temporal consistency}
\label{sec.consistency}

We generate candidate action unit pairs $(u_a, u_b)$ by exploiting the temporal consistency and smoothness properties of the human action evolution. To this end, we use the Gram matrix constructed in the previous step and process it to find the top candidates. We select matched segments which have a similarity score greater than some positive threshold $T$. In our experiments we set $T$ to the one standard deviation above the mean of elements of $G$. If $T$ is non-positive we declare no matching segments. We then find the top temporally consistent candidate matches via a simple search algorithm, which we describe below.

Note that the problem of finding the largest common subsequence in two sequences of length $A$ and $B$ has a computational complexity of $O(A^2B^2)$, which is prohibitively expensive. Furthermore, we are interested in finding the top $K$ best matches, not just the single best match. To alleviate the computational cost, we reduce the search space by employing some heuristics. We formalize the problem of identifying action unit pairs as a bipartite graph matching problem with temporal consistency constraints as show in Figure~\ref{fig:bipartite}. We assume that the graph is sparse (via discarding edges with weight less than our threshold $T$) and that the matched subsequences of video segments have the same length and satisfy the temporal consistency (i.e., are in the same order). For example, the subsequence $\bw_i^a, \ldots, \bw_{i+k}^a$ matches the subsequence $\bw_j^b, \ldots, \bw_{j+k}^b$ if and only if the inner-product $\bw_{i+p}^a \cdot \bw_{j+p}^b \ge T$ for all $p \in \{0, \ldots, k\}$.

Our proposed algorithm is shown in \algref{algo.greedy}. Given sequences of encoded video segments $W_a$ and $W_b$ for videos $X_a$ and $X_b$, respectively, the algorithm finds all runs of matching action unit candidates containing more than $L$ video segments ($L > 1$). Such a strategy allows us to obtain longer action units, i.e., beyond the size of the original window size $l_w$, without resorting to a multiple scale strategy (with multiple window sizes) as commonly done in the supervised action detection literature~\cite{Rohrbach2012,Yeung2016}.

\begin{algorithm}
\small
\SetAlgoLined
\KwInput{Sequence $W_a=\langle \bw_1^a, \ldots, \bw_A^a \rangle$}
\KwInput{Sequence $W_b=\langle \bw_1^b, \ldots, \bw_B^b \rangle$}
\KwInput{Minimum match length $L$ and threshold $T>0$}
\KwOutput{Candidate detections $J$}
Construct gram matrix G as $G_{i,j} = \bw_i^a \cdot \bw_j^b$\;
Initialize candidate detection graph $J$ ($J_{i,j} = 0, \forall i,j$)\;
\For{$i\leftarrow 1$ \KwTo $A-L$}{
\tcp{find all $\bw_j^b$ that match $\bw_i^a$}
$C_0 = \{ j \in \{1, \ldots, B\} \mid G_{i,j} > T\}$\;
\For{$k \leftarrow 1$ \KwTo $L-1$}{
        \tcp{find all $\bw_j^b$ that match $\bw_{i+k}^a$}
	$C_k = \{ j \in \{1, \ldots, B\} \mid  G_{i+k,j} > T\}$\;
        \tcp{remove temporallly inconsistent matches from $C_0$}
        $C_0 = C_0 \cap (C_k - k)$\;
}
\tcp{update candidate detection graph}
\For{$k \in \{1, \ldots, L\}$ \emph{\textbf{and}} $c \in C_0$}{
    $J(i+k\!-\!1,c+k\!-\!1) = G(i+k\!-\!1,c+k\!-\!1)$\;                
  }	
}
\caption{Candidate generation of action unit matches with temporal consistency.}
\label{algo.greedy}
\end{algorithm}

Once we obtain set of candidates we use non-maximum suppression to get rid of the redundant candidates. Two pairs of candidate matching action units are considered redundant if they overlap with more than 0.5 IoU. To be precise, let $u_1^a$ and $u_2^a$ be two sequence of video segments from video $X_a$ and let $u_1^b$ and $u_2^b$ be two sequences of video segments from video $X_b$. Let us assume that pairs $(u_1^a, u_1^b)$ and $(u_2^a, u_2^b)$ are matching. Then these two pairs of matching action units are redundant if $u_1^a$ and $u_2^a$ overlap with IoU greater than 0.5 and $u_1^b$ and $u_2^b$ overlap with IoU greater than 0.5. In such situations we keep only the pair with highest matching score. The matching score of a candidate action unit pair $(u_*^a, u_*^b)$ is the sum of scores of all matched segments (of size $l_w$) within that candidate action unit pair, i.e., $\sum_{\bw^a, \bw^b \in (u_*^a, u_*^b)} \bw^a \cdot \bw^b$.


\subsection{Baselines methods}
\label{sec.baseline}

In this section we present two baseline methods that utilizes rank-pooling for comparison.

\textbf{Clustering method:}
Given each video $X = \left< \bx_1, \bx_2, \ldots, \bx_n \right>$, we first generate video segments as above with length $l_w$ with a stride of one. Then we temporally encode each segment using approximate rank pooling~\cite{Bilen2016} without using any non-linear operations on the input data. Approximate rank-pooling is used due to its computational efficiency (it has constant time complexity). Let us denote the sequence of temporally encoded output vectors by $V = \left< \bv_1, \bv_2, \ldots, \bv_{n-l_w+1} \right>$. To obtain segments with similar dynamics, we cluster each of the sequences $V$ into $k$ clusters using $k$-means. The goal is to find clusters that are, in fact, temporally meaningful and valid temporal segments. To further enforce this, we use the following simple trick. We modify each element of the sequence $V = \left< \bv_1, \bv_2, \ldots \right>$ such that $\bv_t^\text{new} = (\bv_t, \beta t)$. This way, we obtain clusters that are correlated in time as well as being dynamically similar. Any cluster that does not have more than some given number of frames (in our case 60) are pruned. Afterwards, we encode each of the valid clusters with temporal rank pooling using \eqnref{eq.steps.us}. Each of the clusters is now time coherent and results in temporally encoded subsequences which we use as candidate action units.

Given a pair of videos, we create such clusterings, one per video and match the clusters of two videos using temporal rank-pooled encoding of segments. The matching is done using cosine similarity. All pairs of clusters having temporal cosine similarity greater than some threshold (0.2 in our experiments) is considered a matching candidate pair for the final evaluation. We keep the top $k$ such detections as candidates.

\textbf{Rank pooling-based matching:}
In our second baseline, we select a window size $l_w$ and a stride $l_s$ and temporally rank pool each video segment starting from $t=1$ up to $t=n,m$ (i.e., to the end of the video). This result in a sequence of rank pooled features $\left< \bw_1^a, \ldots, \bw_{n-l_w}^a \right>$ and $\left< \bw_1^b, \ldots, \bw_{m-l_w}^b \right>$. As before we construct the gram matrix using using \eqnref{eq:gram}. We then keep only the pairs of matched sub-sequences having cosine similarity greater than some threshold (again, 0.2 in our experiments) as candidate detections. This method is equivalent to our proposed algorithm when the minimum matched sequence length $L$ is set to zero.

%% file: experiments.tex
In this section we report results from extensive experiments on action detection by action matching. We start by outlining our evaluation criteria and datasets used.

\subsection{Evaluation criteria}
Given a pair of videos from the ground truth temporal annotation we obtain the start and end of each action. Each video may contain more than one instance of human action. Therefore, there can be more than one matching pairs of ground truth action units. For example, let us assume that there are $N_a$ ground truth action units from action class $y$ in video $X_a$ and $N_b$ number of ground truth action units from the same action class in video $X_b$. Therefore, there are $N_a \times N_b$ matching pairs of ground truth action units. Then a perfect method would be able to detect all of them within a specific IoU threshold (0.5) as explained in Section~\ref{sec:problem}. To evaluate algorithms we use precision, recall and F1-score. During evaluations, we ignore all redundant candidates (action unit pairs) and keep only a single best candidate (action unit pair) per ground truth pair. This is done only at the evaluation. Candidate generation algorithm has no access to temporal annotations.
For a given pair of videos, and generated candidate pairs of action units, the precision, recall and F1-score is define as follows:
\begin{equation}
 P = \frac{\# \text{ of correct candidate action unit pairs}}{\# \text{of candidates generated}} \times 100
\end{equation}
\begin{equation}
 R = \frac{\# \text{ of correct candidate action unit pairs}}{\# \text{ of ground truth action unit pairs}} \times 100
\end{equation}
\begin{equation}
 F_1 = \frac{2PR}{P + R}
\end{equation}

\subsection{Datasets}


\textbf{THUMOS dataset~\cite{THUMOS15}:}
THUMOS'15 dataset includes four parts: training data, validation data, background data and test data. The training data is based on the UCF101 [2] action dataset, where videos are temporally trimmed (each video usually contains one instance of the action without irrelevant frames). A subset of 20 action classes out of 101 is employed for this task. The training videos are not useful for our task. We use the validation and test set of the THUMOS'15 dataset. Both the validation and the test sets consist of temporal annotations (start and end time) of all instances of the actions occurring in the validation videos. Altogether, there are 412 videos belonging to 20 action classes. Some videos contain multiple human action classes. Some pairs of videos do not contain any matching human action. We ignore any such pairs of videos from the evaluations. There are 6325 temporal annotations in this dataset. We ignore any pair of videos that has only a single action unit (ground truth detection). Finally, we end up with 5094 ground truth detection pairs over all 412 videos.

\textbf{MPII Cooking dataset~\cite{Rohrbach2012}:}
This dataset contains 65 different cooking activities, such as cut slices, pour spice, etc., recorded from 12 participants. In total there are 44 videos with a total length of more than 8 hours or 881,755 frames. The dataset contains a total of 5,609 annotations of 65 activity categories. Since each video is very long we use all possible pairs during the evaluation. Therefore, there are 946 total number of video pairs for evaluation.

\textbf{IKEA dataset:}
This dataset contains 20 sequences of different people assembling the same IKEA drawer. Each sequence consists of approximately 300-400 frames. The viewpoint of the camera for each sequence is approximately the same. The dataset and the annotations we used in our experiments will be available\footnote{\url{http://roboticvision.org/codedata/}}.

\subsection{Feature and frame representations}
For MPII Cooking activities dataset, we use provided dense trajectory features encoded with bag-of-word. We use HOG, HOF, MBH based trajectory features quantized into 4000 visual words as the frame representations~\cite{Rohrbach2012}. 
For THUMOS dataset, we use the residual network features~\cite{He2016} (152-layer network).
For IKEA dataset, we evaluate our approach using similar features to the MPII Cooking activities dataset in addition to the 152-layer residual network features.
We use publicly available rank pooling code\footnote{\url{https://bitbucket.org/bfernando/videodarwin}}.

\textbf{Baseline 1 details:} We apply approximate rank pooling on input sequences with window size ($l_w$) of 61 (21 for IKEA dataset) and stride 1. Afterwards, we L2 normalize the temporally pooled sequences. We cluster each sequence into 10 clusters. Experimentally we found that 10 clusters is reasonable as it generates roughly 100 candidate action detections. Before clustering step, each temporal pooled vector at time t is concatenated with the time variable such that the new vector $\bv_{t}^\text{new} = (\bv_t,  \beta t)$ where $\beta = 0.001$. Then we keep candidate temporal segments (clusters) if they are temporally consistent and longer than 60 frames (20 frames for IKEA dataset). Given a pair of videos we have $n$ temporal segments for the first video and $m$ segments for the second and find candidates as explained in section~\secref{sec.baseline}.

\subsection{Results}
First, we report results for two baseline methods and proposed effective temporal consistency method using Cooking activities dataset in Table~\ref{tabel.cookingresults}, IKEA dataset in Table~\ref{ikea.results} and THUMOS15 dataset in Table~\ref{tabel.thumosresults}. For Cooking activities and THUMOS15 datasets, we use window sizes of 61 (stride 10) and minimum match length $L$ of size $L=10$, and use top 100 candidate detections for evaluation. 

Results in Table~\ref{tabel.cookingresults} suggest that the best individual feature is MBH (F1 score of 14.1). Interestingly, the second best feature is the HOG feature. Most interestingly, the temporal consistency method improves over other two baseline by significant margin in terms of F1-score. For MBH features, the clustering method obtains F1 score of 4.4, rank pooling based matching obtains 4.8 while the temporal consistency method improves results to \textbf{14.1}. Similar trends can be observed for both IKEA dataset and the challenging THUMOS15 dataset. The temporal consistency method outperforms other two baselines over all three datasets using both trajectory features as well as deep residual network features. In some instances, for MPII Cooking activities dataset, the improvement of the temporal consistency method is more than twice other methods in terms of F1 score. Interestingly, for IKEA dataset the rank pooling matching obtains significant precision values. Perhaps this is because this is a relatively small dataset. 

We also report results by fusing the HOG, HOF and MBH features. HOG, HOF and MBH features are fused using the average gram metric (similar to average Kernel) for Rank pooling-based matching and temporal consistency methods. Other fusion methods such as merging the candidates from different features did not improve our results perhaps because such a strategy would not be able to exploit the advantages of temporal consistency. For clustering method, we use early fusion (concatenation of feature vectors). Even after the fusion, we see that temporal consistency method is effective. Therefore, we conclude that rank pooling based temporal consistency method is useful for unsupervised action detection by action matching task.

\begin{table}[t]
\scriptsize
\begin{center}
\begin{tabular}{|l|c|c|c|}
\hline
Method & Rc. (\%) & Pr. (\%) & F1 (\%) \\
\hline\hline
\multicolumn{4}{|c|}{HOG} \\ \hline
Cluster method 			        & \textbf{17.4} & 4.1 & 6.6 \\ 
Rank pooling-based matching     & 3.8  & 3.1 & 3.4 \\
Temporal consistency 		    & 14.1 & \textbf{13.7} & \textbf{13.9} \\ 
\hline
\multicolumn{4}{|c|}{HOF} \\ \hline
Cluster method 			        & 7.2 & 2.2 & 3.4 \\ 
Rank pooling-based matching 	& 3.2  & 2.8 & 3.0 \\ 
Temporal consistency 		    & \textbf{9.5} & \textbf{9.6} & \textbf{9.6} \\
\hline
\multicolumn{4}{|c|}{MBH} \\ \hline
Cluster method 			        & 10.7 & 2.8 & 4.4 \\ 
Rank pooling-based matching 	& 5.3  & 4.4 & 4.8 \\
Temporal consistency 		    & \textbf{14.2} & \textbf{14.0} & \textbf{14.1}  \\
\hline
\multicolumn{4}{|c|}{Fused} \\ \hline
Cluster method & \textbf{13.1}& 5.4 & 7.6 \\
Rank pooling-based matching & 7.2  & 6.4 & 6.8 \\
Temporal consistency & 11.7 & \textbf{21.6} & \textbf{15.1}  \\ 
\hline
\end{tabular}
\end{center}
\caption{Unsupervised action detection results for MPII Cooking dataset.}
\label{tabel.cookingresults}
\end{table}

\begin{table}[t]
\scriptsize
\begin{center}
\begin{tabular}{|l|c|c|c|}
\hline
Method & Rc. (\%) & Pr. (\%) & F1 (\%) \\
\hline\hline

\multicolumn{4}{|c|}{HOG} \\ \hline
Cluster method 			&2.2 &2.4  & 2.1 \\
Rank pooling-based matching 	&2.1 &\textbf{57.4} &4.0 \\
Temporal consistency 		&\textbf{14.2} &17.5 &\textbf{15.7} \\ 
\hline
\multicolumn{4}{|c|}{HOF} \\ \hline
Cluster method 			&3.0 & 5.1 & 3.6 \\
Rank pooling-based matching 	&2.1 &\textbf{58.2}  & 4.0 \\
Temporal consistency 		&\textbf{22.5}	&24.9 	& \textbf{23.6} \\ 
\hline
\multicolumn{4}{|c|}{MBH} \\ \hline
Cluster method 			&0.9 &1.0 &0.9\\
Rank pooling-based matching 	&2.1 &\textbf{58.1} & 4.0 \\
Temporal consistency 		&\textbf{17.8} &22.9 &\textbf{20.0} \\ 
\hline
\multicolumn{4}{|c|}{Deep} \\ \hline
Cluster method &0.15 &0.13 &0.14 \\
Rank pooling-based matching &2.7 &\textbf{51.1} &5.0 \\
Temporal consistency &\textbf{19.6} &24.6 &\textbf{21.1} \\ 
\hline
\multicolumn{4}{|c|}{Fused} \\ \hline
Cluster method &3.8 &3.6 &3.5 \\
Rank pooling-based matching &2.1 &\textbf{58.3} &4.1 \\
Temporal consistency &\textbf{25.3} &24.9 &\textbf{25.1} \\ 
\hline
\end{tabular}
\end{center}
\caption{Unsupervised action detection results for IKEA dataset.}
\label{ikea.results}
\end{table}

\begin{table}[t!]
\scriptsize
\begin{center}
\begin{tabular}{|l|c|c|c|}
\hline
Method & Rc. (\%) & Pr. (\%) & F1 (\%) \\
\hline\hline
Cluster method 			& 12.5 & \textbf{17.6} & 14.6\\
Rank pooling-based matching 	& 11.5 & 4.8 &	 6.7 \\
Temporal consistency 		& \textbf{24.2}	& 16.3 	& \textbf{19.5}\\  
\hline
\end{tabular}
\end{center}
\caption{Unsupervised action detection results for THUMOS15 dataset.}
\label{tabel.thumosresults}
\end{table}

\subsection{Impact of ARMA model}

Next we evaluate the impact of ARMA model on unsupervised action detection task using all three datasets. 
We only report the F1 score for clarity. 
Results are reported in Table~\ref{tbl.arma}. 
Results suggest that ARMA smoothing process obtains better results compared to the Time Varying Mean (TVM)~\cite{Fernando2016} except for the HOG features in the Cooking activities dataset. 
The impact of ARMA model over THUMOS15 dataset is about 1.7\%. 
We conclude that ARMA model is better suited than the TVM for unsupervised action detection task using rank-pooling and temporal consistency algorithm.

\begin{table}[t]
\scriptsize
\begin{center}
\begin{tabular}{|l|c|c|c|c|c|}
\hline
Method & HOG & HOF & MBH &  Resnet \\ \hline\hline
IKEA - TVM  &15.7  &23.6 &20.0  &21.1  \\ 
IKEA - ARMA &\textbf{18.4}  &\textbf{26.5} &\textbf{21.9}   &\textbf{21.8} \\ \hline

Cooking - TVM  & \textbf{13.9} & 9.6 & 14.1 &  -- \\ 
Cooking - ARMA & 13.7 & \textbf{10.3} & \textbf{14.2} & -- \\ \hline

THUMOS15 - TVM  & -- & -- & --  &  19.5 \\ 
THUMOS15 - ARMA & -- & --  & -- & \textbf{21.2} \\ \hline
\end{tabular}
\end{center}
\caption{Impact of ARMA model on unsupervised action detection results.}
\label{tbl.arma}
\end{table}

\begin{figure*}[t]
 
\begin{center}
\resizebox{3\totalheight}{!}{
  \subfloat[][]{\includegraphics[width=0.28\linewidth]{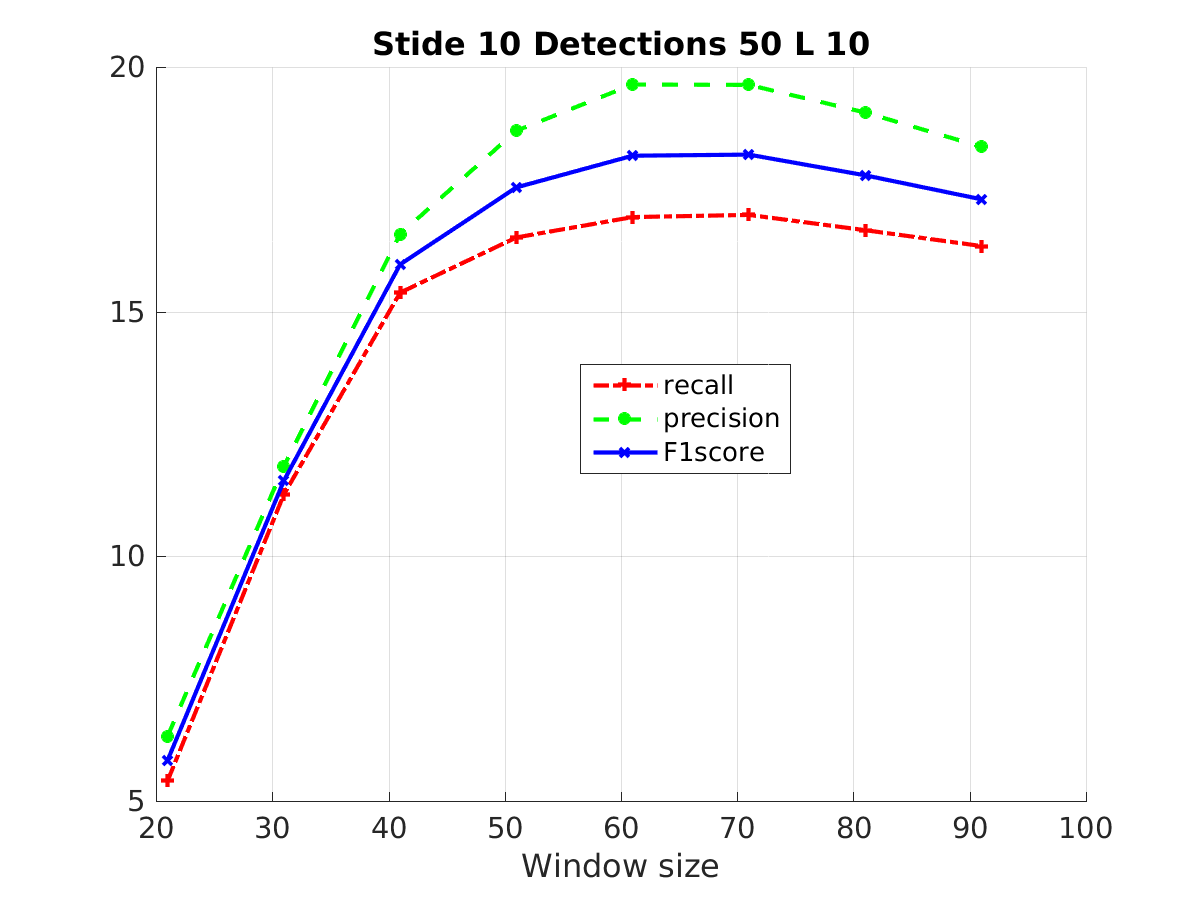}}   
  \subfloat[][]{\includegraphics[width=0.3\linewidth]{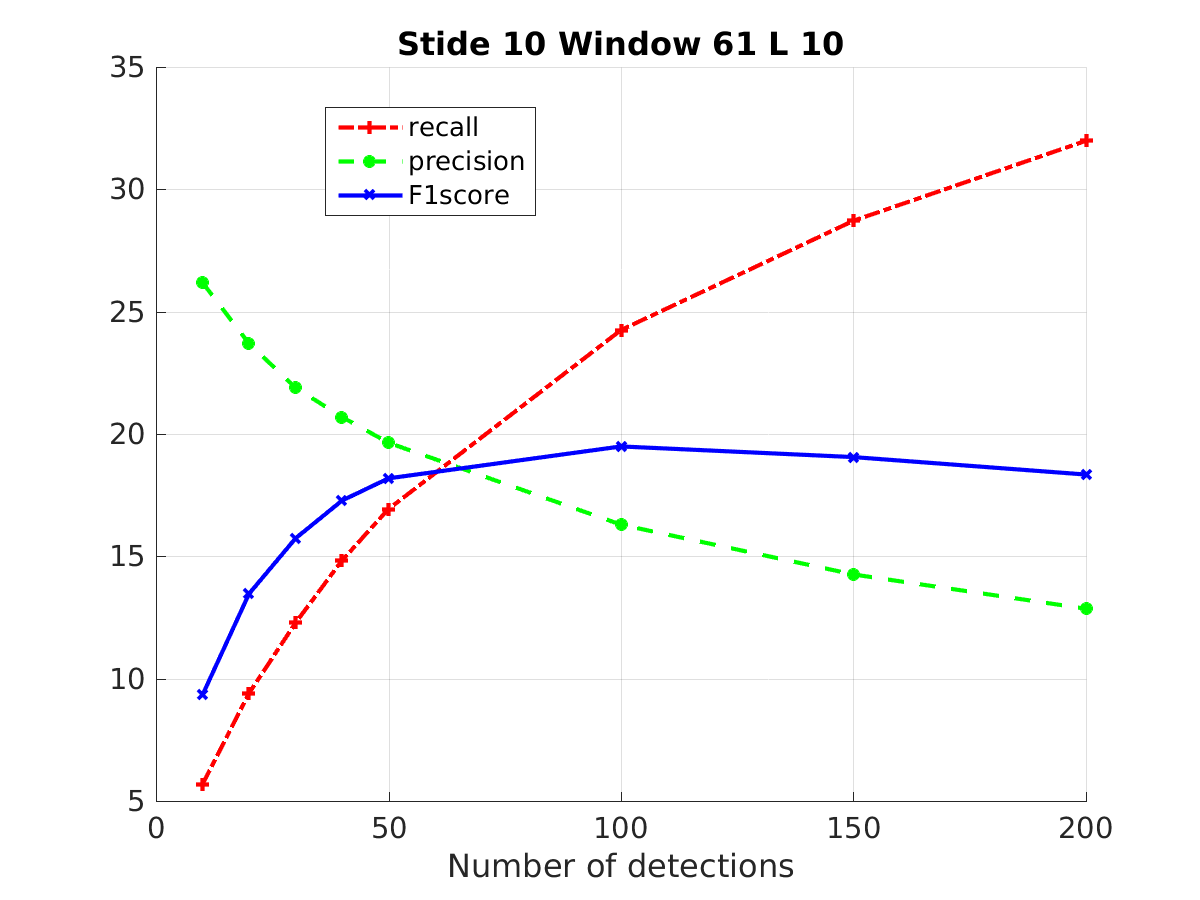}}
  \subfloat[][]{\includegraphics[width=0.3\linewidth]{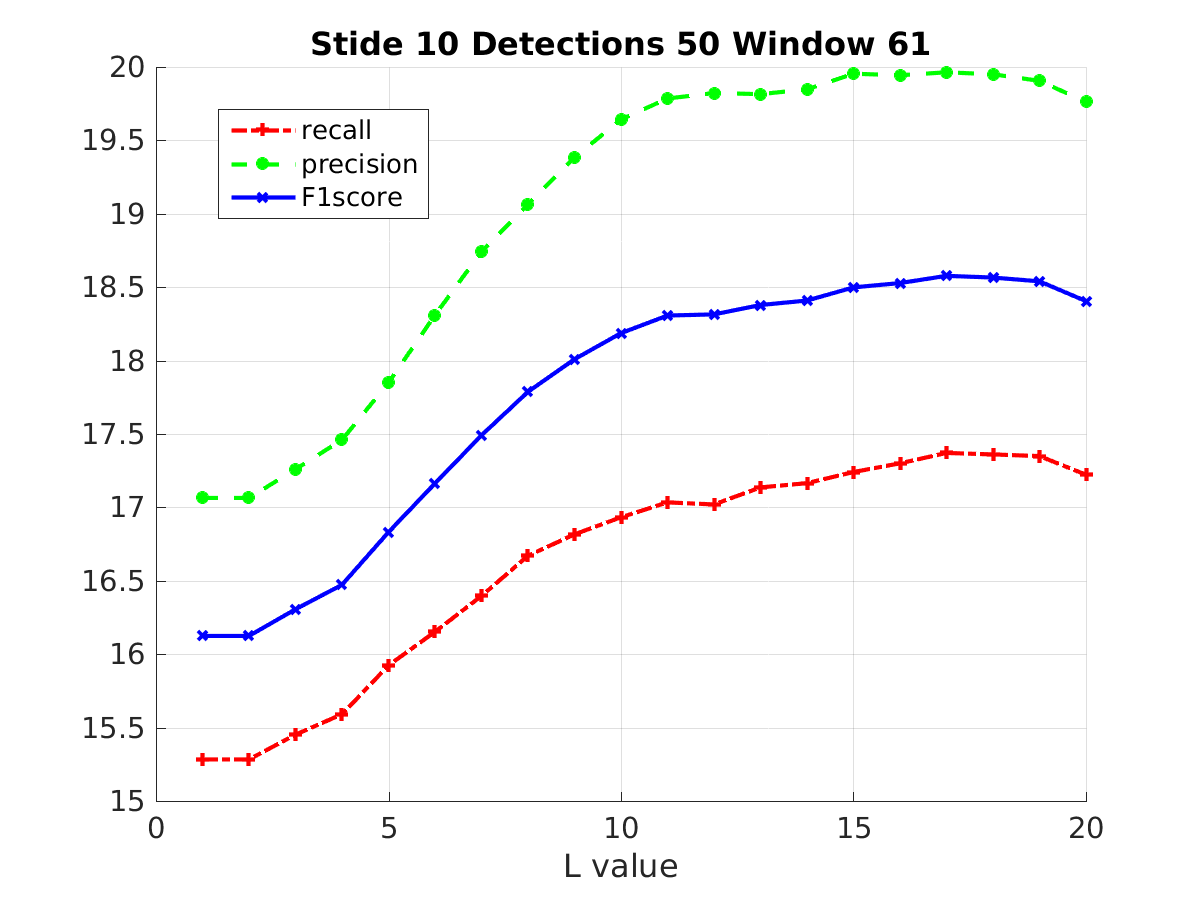}}
  }
\end{center}
\caption{Effect of several parameter on temporal consistency method. (a) window size, (b) number of candidate detections and (c) maximum temporal segment length (L) is evaluated using THUMOS15 dataset.}
\label{fig:EffectOfParamsTHUMOS}
\end{figure*}

\begin{figure}[t]
\begin{center}
  \includegraphics[width=0.99\linewidth]{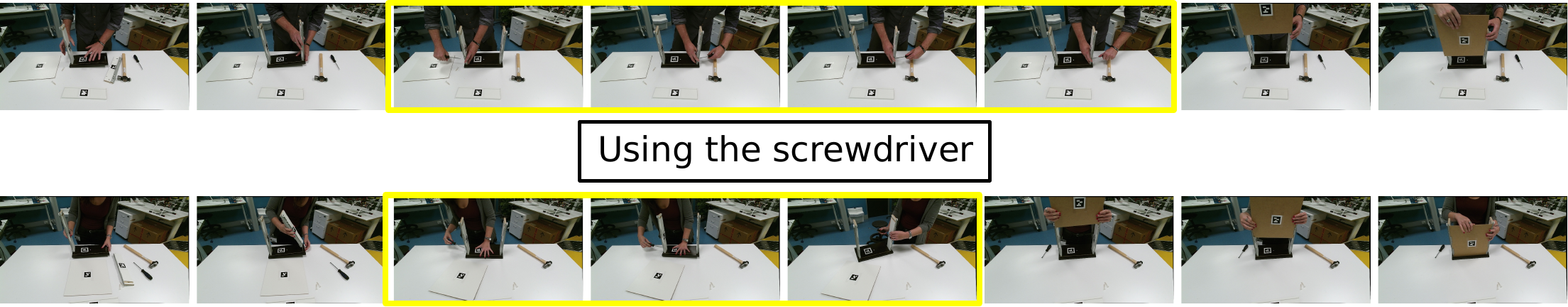}
\end{center}
\caption{Visualizing unsupervised action detection from IKEA dataset}
\label{fig:IkeaSamplevisual}
\end{figure}

\begin{table}
\scriptsize
\begin{center}
\begin{tabular}{|l|c|c|c|}
\hline
Class & Recalled \% of action segments \\ \hline \hline
GolfSwing   &   43.0 \\\hline
BaseballPitch   &   35.2 \\\hline
CleanAndJerk   &   33.1 \\\hline
SoccerPenalty   &   25.8 \\\hline
Shotput   &   22.2 \\\hline
\end{tabular}
\end{center}
\caption{Class based action discovery performance for THUMOS dataset.}
\label{tabel.thumo.action.discovery}
\end{table}

\begin{table}
\scriptsize
\begin{center}
\begin{tabular}{|l|c|c|c|}
\hline
Class & Recalled \% of action segments \\ \hline \hline
take-out-from-drawer & 57.0 \\ \hline 
take-out-from-cupboard & 47.7 \\ \hline 
take-out-from-fridge & 34.3 \\ \hline 
take-\&-put-in-cupboard & 20.0 \\ \hline 
take-\&-put-in-fridge & 16.7 \\ \hline 
\end{tabular}
\end{center}
\caption{Class based action discovery performance for Cooking dataset (Top 5 classes)}
\label{tabel.cooking.action.discovery}
\end{table}

\subsection{Evaluating temporal consistency parameters}
In this section we evaluate the effectiveness of several parameters of the temporal consistency method. We use the THUMOS15 dataset for parameter evaluation. For this experiment we use TVM method (not ARMA). We evaluate the impact of window size $l_w$, number of candidate detections, and minimum matched sequence length $L$. Results are shown in Figure~\ref{fig:EffectOfParamsTHUMOS}. 
As it can be seen from Figure~\ref{fig:EffectOfParamsTHUMOS} (a), the results improve with bigger window size but results start to decrease after window size of 61. The results in the second plot (Figure~\ref{fig:EffectOfParamsTHUMOS} (b)) is not surprising. It indicates as the number of detections increases, the precision drops while the recall improves. 
However, the best F1 score is obtained for 60 detections using a window size of 60 and L value of 10. Next in Figure~\ref{fig:EffectOfParamsTHUMOS} (c), we see that as the minimum temporal candidate length $L$ increases, the results improve significantly up to about $L$ value of 19. 
This plot is very interesting and suggests that relatively large temporal candidates are better suited. This is a clear advantage of our rank pooling-based temporal consistency algorithm. Note that our temporal consistency method can be applied over other temporal encoding methods as well (not just limited to rank pooling). The main advantage of our temporal consistency method is that even if one uses a fixed window size, yet able to obtain variable length action unit candidates beyond the window size. Results suggest that such a strategy allows us to improve the detection performance without additional temporal encodings of different window sizes. We conclude that our temporal consistency method is very useful for unsupervised action detection task.

\subsection{Class-based analysis and action discovery}
In this section we compute the class-based analysis on detected action segments. In the following we show the recall (percentage of discovered action segments) of each action category using our temporal consistency algorithm. We report the best 5 classes for Cooking activities and THUMOS15 datasets in Table~\ref{tabel.thumo.action.discovery} and Table~\ref{tabel.cooking.action.discovery} respectively. 
Some of the detected human actions for IKEA dataset is also shown in Figure~\ref{fig:IkeaSamplevisual}.
Some aligned videos from MPII-Cooking dataset can be found here\footnote{\url{https://www.youtube.com/watch?v=_84QH2nxGRY}}.
We conclude that the top detected action segments from our method are class specific and can be used as an action discovery technique.

%% file: conclusion.tex
We have proposed a novel task called unsupervised action detection by action matching. In this task the objective is to find pairs of video segments that share a common human action from a long pair of videos. It is an unsupervised task as the task is agnostic to the action class which make it useful for many real world applications. We have presented an effective and efficient method for discovering such human action pairs. We exploit the temporal consistency and the temporal evolution of videos to discover such pairs of video segments. We obtained promising results on three action detection datasets including MPII Cooking activities dataset and THUMOS15 challenge dataset. 
We believe in future the proposed task would be evolved to jointly learn video representations, action categories and action detectors in an unsupervised or semi-supervised manner while solving many real world problems.